\documentclass[11pt]{article}
\usepackage{coling2020}
\usepackage{times}
\usepackage{url}
\usepackage{latexsym}
\usepackage{graphicx}
\usepackage{subfig}
\usepackage{lipsum}
\usepackage{color}
\usepackage{booktabs}

\colingfinalcopy 

\title{Enabling Interactive Transcription in an Indigenous Community}

\author{
  \'Eric Le Ferrand,$^{1,2}$ Steven Bird,$^1$ and Laurent Besacier$^2$ \\
  $^1$Northern Institute, Charles Darwin University, Australia \\
  $^2$Laboratoire Informatique de Grenoble, Université Grenobles Alpes, France
  }

\date{}

\begin{document}
\maketitle
\begin{abstract}
We propose a novel transcription workflow which combines spoken term detection and human-in-the-loop,
together with a pilot experiment.
This work is grounded in an almost zero-resource scenario where only a few terms have so far been identified,
involving two endangered languages.
We show that in the early stages of transcription, when the available data is insufficient to train a robust ASR system,
it is possible to take advantage of the transcription of a small number of isolated words in order to bootstrap the transcription of a speech collection.
\end{abstract}

\section{Introduction}
\label{intro}
\blfootnote{
    %
    %
    %
    %
    \hspace{-0.65cm}  
    This work is licensed under a Creative Commons 
    Attribution 4.0 International Licence.
    Licence details:
    \url{http://creativecommons.org/licenses/by/4.0/}.
    %
    %
}
In remote Aboriginal communities in Australia, many efforts are made to document traditional knowledge including rock art, medicinal plants, and food practices.
While it may be relatively straightforward to capture spoken content, transcription is time-consuming and has been described as a bottleneck \cite{brinckmann2009transcription}.
Transcribing is often seen as an obligatory step, to facilitate access to audio.
Efforts have been made to speed up this process using speech recognition systems, but the amount of data available in Indigenous language contexts is usually too limited for such methods to be effective.


Recent research has shown the efficacy of spoken term detection methods when data are scarce \cite{menon2018fast,Menon2018}. Taking advantage of the transcription of a few words would allow us to propagate it through the speech collection and thus assist language workers in their regular transcription work.
So-called ``sparse transcription'' would be also a way to navigate a speech collection and allow us 
 to be selective about what needs to be transcribed \cite{Bird20}.
Several tools exist for manual transcription, such as Elan and Praat \cite{wittenburg2006elan,boersma1996praat}.
However such transcriptions are often made in isolation from the speech community \cite{australia2014angkety},
and so we miss out on the opportunity to take advantage of the interests and skills of local people
to shape and carry out the transcription work.

We present a fieldwork pipeline which combines automatic speech processing and human expertise to support speech transcription in almost-zero resource settings.
After giving the background, we detail the workflow and propose a pilot experiment on two very low-resource corpora.


\section{Background}

Existing approaches to automatic transcription of endangered languages involve methods that
have been developed for automatic speech recognition. While a few hours of transcribed speech can be enough to train single-speaker models \cite{gupta2020speech}, speaker-independent models require a large amount of training data to produce useful transcriptions \cite{gupta2020automatic,foley2018building}. 
Moreover, they draw language workers and speakers into the time-consuming task of exhaustive transcription,
forcing them to transcribe densely, including passages that may be difficult or impossible
given the early state of our knowledge of the language.
A more suitable approach, we believe, involves beginning with stretches of speech where we have the greatest confidence,
and only later tackling the more difficult parts.



Spoken term detection involves retrieving a segment in a speech collection, given an example.
With this method, it is possible to sparsely transcribe the corpus,
i.e., take advantage of an existing transcription (or a list of spoken words) and identify tokens throughout the collection \cite{Bird20}.

Dynamic Time Warping (DTW) \cite{sakoe1978dynamic} and its more advanced versions \cite{park2005towards,jansen2011efficient} can be used to match a query in a speech collection, but they can be computationally  expensive.
DTW ``aligns two sequences of feature vectors by warping their time axes to achieve an optimal match'' \cite{menon2018fast}.
A common method is to compute DTW, sliding the spoken query over each utterance in the corpus.
In addition, several speech representations have been considered in the literature for the task of word-spotting. 
\newcite{kamper2015unsupervised} and \newcite{menon2019feature} explore feature extraction from autoencoders (AE) and correspondence autoencoders (cAE) and show how  top-down constraints can improve the quality of the hidden representation for spoken term detection tasks.
\newcite{schneider2019wav2vec} introduce wav2vec, a self-supervised model for speech representation learning which is based on contrastive predictive coding, and apply this to supervised ASR.

\section{Proposed Workflow}

\subsection{Interactive and Sparse Transcription}
\label{sec:sparse}

\begin{figure*}[!b]
    \begin{minipage}{\linewidth}
	\begin{minipage}{.35\linewidth}
		\centering
		\subfloat[Starting point]
		{\label{setup:0}\framebox{\includegraphics[width=0.85\textwidth]{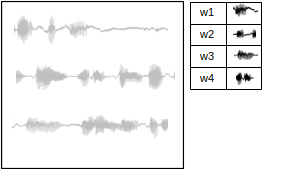}}}
	\end{minipage}
	\hspace{-0.55cm}
	\begin{minipage}{.35\linewidth}
		\centering
		\subfloat[Spoken term detection]
		{\label{setup:1}\framebox{\includegraphics[width=0.85\textwidth]{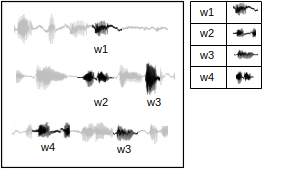}}}
	\end{minipage}
	\hspace{-0.55cm}
	\begin{minipage}{.35\linewidth}
		\centering
		\subfloat[Manual confirmation]
		{\label{setup:2}\framebox{\includegraphics[width=0.85\textwidth]{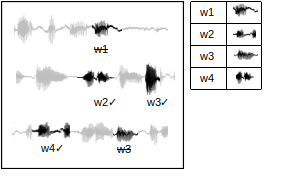}}}
	\end{minipage}
	\end{minipage}\\[2ex]
	\begin{minipage}{\linewidth}
	\begin{minipage}{.35\linewidth}
		\centering
		\subfloat[Confirmed hits]
		{\label{setup:3}\framebox{\includegraphics[width=0.85\textwidth]{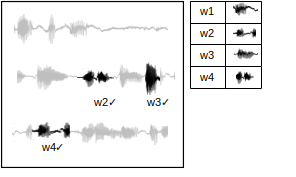}}}
	\end{minipage}
	\hspace{-0.55cm}
	\begin{minipage}{.35\linewidth}
		\centering
		\subfloat[Extra examples added to lexicon]
		{\label{setup:4}\framebox{\includegraphics[width=0.85\textwidth]{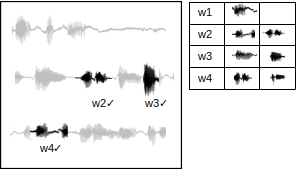}}}
	\end{minipage}
	\hspace{-0.55cm}
	\begin{minipage}{.35\linewidth}
		\centering
		\subfloat[Expanded lexicon]
		{\label{setup:5}\framebox{\includegraphics[width=0.85\textwidth]{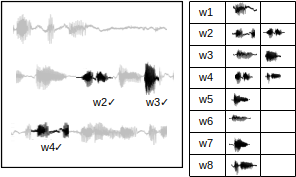}}}
	\end{minipage}
	\end{minipage}
	\caption{One iteration of our Interactive Sparse Transcription Workflow}
	\label{fig:sparse_workflow}
\end{figure*}


The key idea of sparse transcription is to use spoken term detection methods to sparsely transcribe a speech collection,
beginning with a small collection of spoken terms.
A spoken term is defined as a chunk of speech considered meaningful by a speaker,
which may be a morph, a word, or a multiword expression.
This collection of terms can be a list of keywords (or morphs) recorded in isolation by a speaker.
Equally, it can be obtained by extracting audio clips from a speech collection.

We set up the workflow as shown in Figure~\ref{fig:sparse_workflow}.
We begin with a lexicon of size \textit{s} and a speech collection (Fig.~\ref{setup:0}).
This lexicon is composed of speech terms 
and their orthographic transcriptions $w_1\ldots w_n$.
We use a spoken term detection method to retrieve speech terms in the collection that match those in the lexicon (Fig.~\ref{setup:1}).
The system presents the $n$ most confidently identified terms and presents them for verification  (Fig.~\ref{setup:2}).
False positives are corrected, i.e., erased from the transcription (Fig.~\ref{setup:3}).
We clip out from the utterances the speech terms correctly retrieved and add them to the lexicon as extra samples of a given entry (Fig.~\ref{setup:4}).
We allow a single entry to have maximum of \textit{m} extra examples. We manually collect new speech units with a speaker, add them to the lexicon (Fig.~\ref{setup:5}) and start a new iteration (Fig. \ref{setup:1}).
We apply this for \textit{i} iterations with a lexicon growing each time.
The size \textit{s} of the lexicon, \textit{n} number of words checked, maximum
\textit{m} of extra examples for each word, and number \textit{i} of iterations are hyperparameters which vary according to the contingencies of the fieldwork. 

\section{Pilot experiment: simulating this interactive scenario}
\label{expe}
\textbf {Speech data.} We apply the pipeline to two corpora.
The first one is a 4h30m corpus in Mboshi\footnote{https://github.com/besacier/mboshi-french-parallel-corpus} \cite{godard2018very}, a Bantu language spoken in Congo Brazaville (ISO mdw).
It consists of 5,130 utterances sentence and word-aligned with an orthographic transcription.
The utterances are elicited from text.
This corpus contains only three speakers, of which one is responsible for 70\% of the corpus.
The second corpus is a very small (0h20m) corpus in Kunwinjku, an Australian Aboriginal language (ISO gup). 
It consists of 301 utterances aligned with an orthographic transcription.
A forced alignment at the word-level has been created using the MAUS aligner \cite{kisler2017multilingual}. The corpus contains 4 guided tours of the same town and one guided tour of another Aboriginal site.
Each tour has been produced by a different speaker. 
To create initial and expanded lexicons, we select the 100 and 60 most frequent words bigger than 3 syllables for Mboshi and Kunwinjku respectively, in order to avoid words which are too short. A speech occurrence of each entry is extracted from the  speech collection using the word-level alignments.

\textbf{Acoustic features.}
In this interactive process, we explore several speech representations to identify those that are most suited to the pipeline,
namely mel-frequency cepstral coefficients (MFCC) and
perceptual linear prediction (PLP) features.
We also use the hidden representation of an auto-encoder (AE) and correspondence auto-encoder (cAE).
For this we use architecture of \cite{kamper2015unsupervised} to self-train a 5-layer stacked AE (instead of 8) on 4h of Mboshi and 2h of Kunwinjku (YouTube videos).
The cAE is trained using similar segments extracted from the speech collections with an unsupervised term discovery tool \cite{jansen2011efficient}.
Finally we also use wav2vec representations \cite{schneider2019wav2vec}.
The wav2vec model is either trained from scratch on Mboshi and Kunwinjku (w2v\_mb and w2v\_kun) or it is adapted from an original English model (w2v\_en\_mb and w2v\_en\_kun).
We experiment with these features, with and without mean and variance normalisation (MVN) \cite{strand2004cepstral}. 

\textbf{Sparse transcription experiments.}
The  workflow described in section \ref{sec:sparse} is applied for 5 iterations for Mboshi and 3 iterations for Kunwinjku with an initial lexicon consisting of 20 words.
20 new words are added at the lexicon at each iteration. We end up with a 100 word lexicon for Mboshi and 60 words for Kunwinjku. The 10 best hits per word are checked ($n=10$) and a maximum of 5 extra examples per word ($m=5$) are added to the lexicon for both corpora.
In addition, in order to avoid unnecessary verification in case the DTW score is too low, the worst score of the correct words checked during the first iteration is used as a threshold for the following iteration. To simulate human verification (Fig.~\ref{setup:2}) we directly compare the hits output by the system with the gold transcriptions of the corpora.


\section{Results: impact of speech feature representations in the workflow}
\label{results}

\begin{figure}[!ht]%
    \centering
    \subfloat[]{{\includegraphics[width=0.5\textwidth]{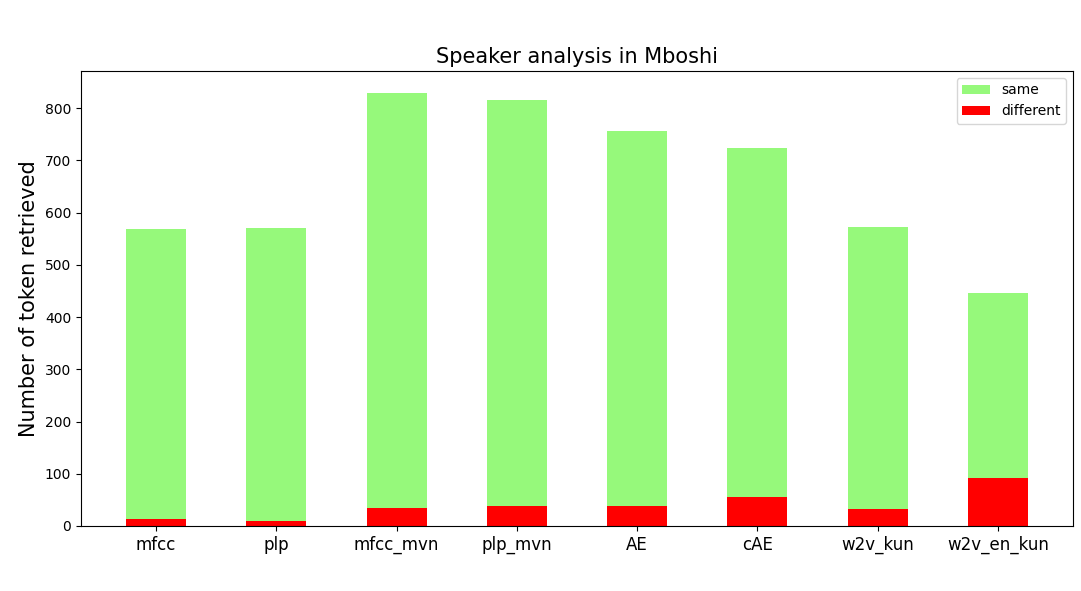} }}%
    \hspace{-5mm}
    \subfloat[]{{\includegraphics[width=0.5\textwidth]{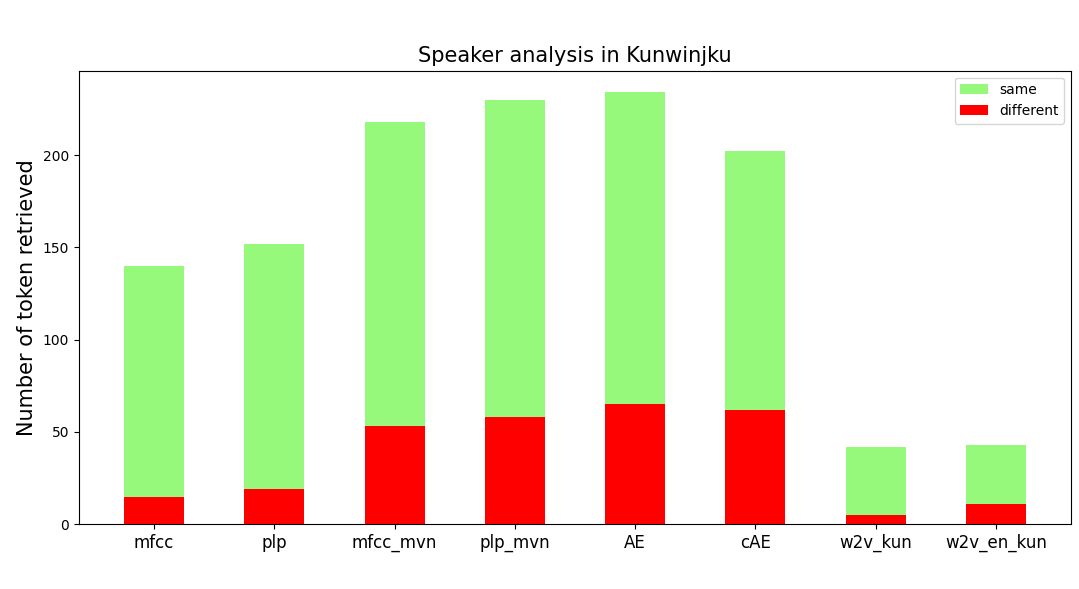} }}%
    \caption{Proportion of same-speaker/different-speaker retrieval for each representation}%
    \label{fig:speaker}%
\end{figure}

We report the results of this new workflow in Table~\ref{table:results}.
The average precision scores (AP) correspond to the mean of the precision of the workflow computed at each iteration. 
The final recall is defined as the number of items retrieved from the full corpus $X$ out of the number of all retrievable items in the corpus, i.e., the intersection of the lexicon $L$ and the corpus $C$, $X/(L \cap C)$.
In other words, this recall corresponds to the coverage of the lexicon related to its tokens in the speech collection.
The impact of MVN is detailed only for the basic representations (PLP and MFCCs) since  normalisation did not show any major influence on the neural representations (AE, cAE and w2v). 
We can make the following comments from the results shown in Table \ref{table:results}:
(a)~mean and variance normalisation (MVN) is important to improve results of basic (PLP and MFCC) features.
Figure~\ref{fig:speaker} shows that normalisation improves retrieval when the query term and the search term are pronounced by different speakers;
(b)~neural AE and cAE features are  normalized-by-design but do not lead necessarily to better performance than PLP and MFCC;
and (c) representations provided by wav2vec are not efficient. The small size of the corpora might be the main obstacle when training efficient self-supervised models for learning speech representations.
%

\begin{table}[!t]

\begin{tabular}{l|l|l|l}
\toprule
Features    & MVN & AP    & final recall            \\ \midrule
mfcc        & no  & 23.87 & 16.78                   \\
mfcc        & yes & \textbf{32.67} & \textbf{23.37} \\
plp         & no  & 23.89 & 16.86                   \\
plp         & yes & 31.64 & 23.03                   \\
w2v\_mb     & no  & 24.57 & 16.88                   \\
w2v\_en\_mb & no  & 19.23 & 13.72                   \\
AE          & no  & 31.93 & 21.51                   \\
cAE         & no  & 27.31 & 20.70                   \\ \bottomrule
\end{tabular}
\hspace{1cm}
\begin{tabular}{l|l|l|l}
\toprule
Features    & MVN & AP    & final recall   \\ \midrule
mfcc        & no  & 15.42 & 30.82          \\
mfcc        & yes & 20.82 & 42.84          \\
plp         & no  & 15.21 & 32.67          \\
plp         & yes & \textbf{22.55} & 44.89 \\
w2v\_kun     & no  & 5.39 & 15.72          \\
w2v\_en\_kun & no  & 5.45 & 15.87          \\
AE          & no  & 22.07 & \textbf{45.30} \\
cAE         & no  & 21.88 & 40.37          \\ \bottomrule
\end{tabular}

\caption{Results for the Mboshi corpus (left) and for the Kunwinjku corpus (right)}
\label{table:results}
\end{table}

Speaker diversity in the two corpora (Sec.~\ref{expe}) has an impact on the final results. 
For Mboshi 70\% of the words to be retrieved in the speech collection are pronounced by the same speaker in the final 
lexicon 
and 37\% for Kunwinjku.
Figure \ref{fig:speaker} reports the proportion of tokens retrieved between same speakers and different speakers over the total number of tokens retrieved for Mboshi and Kunwinjku.
The first observation is that the pipeline mostly retrieves terms pronounced by the 
speaker. It is clear that MVN improves performance for both same-speaker and different-speaker retrieval. 
We also note that, while wav2vec representation is overall less performant, it seems promising for extracting more speaker-independent representations, as illustrated by the results obtained with w2v\_en\_mb features. Future work will investigate this by leveraging more raw speech in Mboshi and Kunwinjku for training better self-supervised (wav2vec) models.

\begin{table}[!b]
    \centering
\begin{tabular}{l|l|l|l}
\toprule
Query   & False Positive & query translation    & hit translation             \\ \midrule
nahne   & mahne          & it (pronoun)         & this (demonstrative)         \\
nemekke & namekke        & that one             & that one (other spelling)   \\
balanda & balanda-ken    & white man            & of the white man (genitive) \\
bininj  & bininj-beh     & man                  & from the man                \\
nemekke & yekke          & that one             & dry season                  \\
mahni   & mahne          & this (demonstrative) & this (other spelling)       \\
\bottomrule
\end{tabular}
\caption{Top false positive generated by the spoken term detection system in Kunwinjku}
\label{error}
\end{table}
Regarding false positives (Table~\ref{error}), the first errors we can observe are different spellings of words (in the gold transcription) referring to the same meaning (e.g., \textit{namekke / nemekke} ``that one'').
Since Kunwinjku is primarily a spoken language, variable spellings are common.
Moreover, because of the morphological complexity of the language, many of the top false positives are actually inflectional variants of the query term (e.g., \textit{balanda / balanda-ken} ``white man / from the white man'').
Allowing matching at a smaller granularity could be a way to achieve wider coverage of the speech collection.


\section{Deployment}

\begin{figure}[!t]%
    \centering
    \subfloat[screenshot of the app]{{\includegraphics[width=0.5\textwidth]{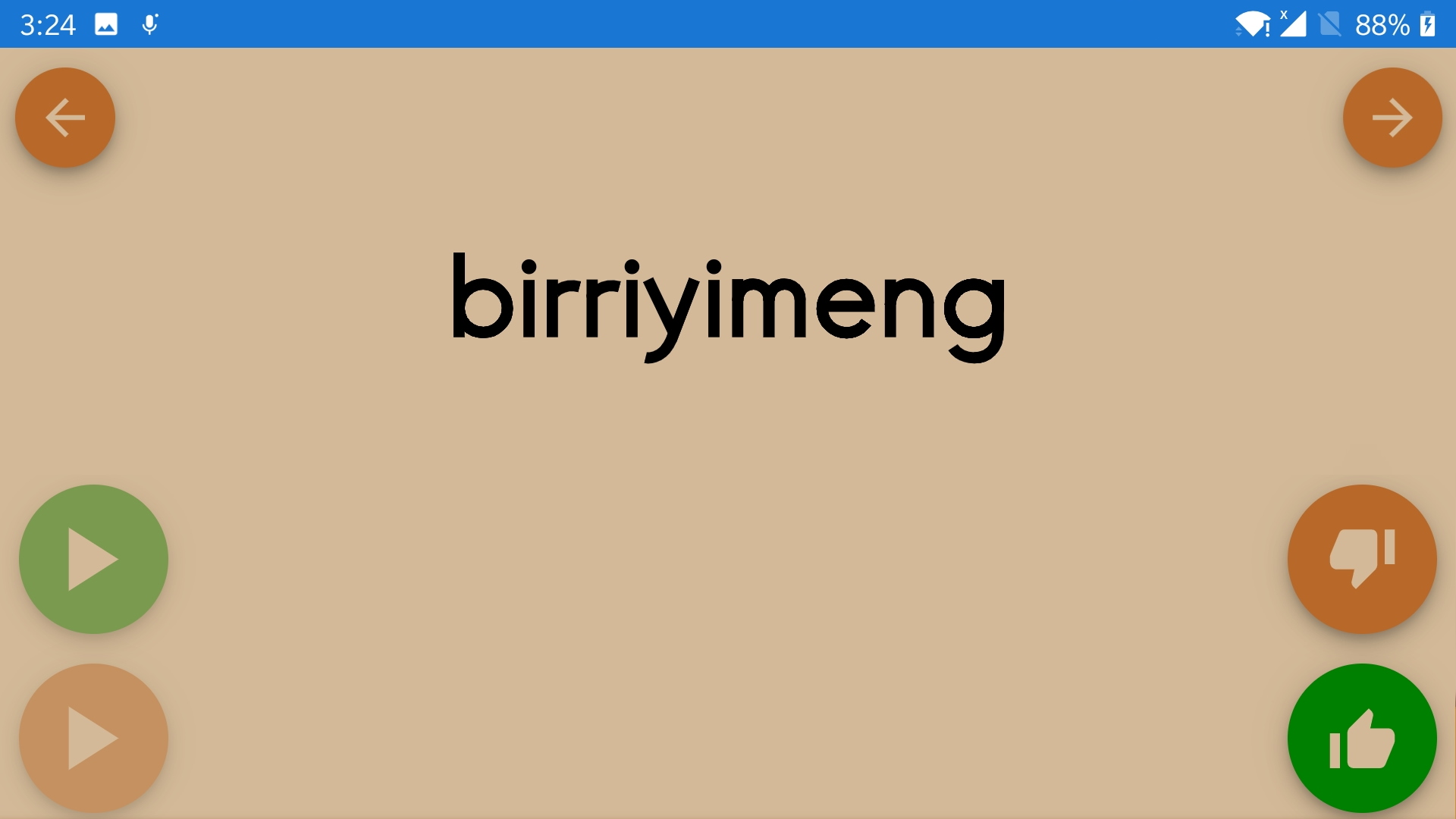} }}%
    \hspace{1.5cm}
    \subfloat[deployment of the app]{{\includegraphics[width=0.36\textwidth]{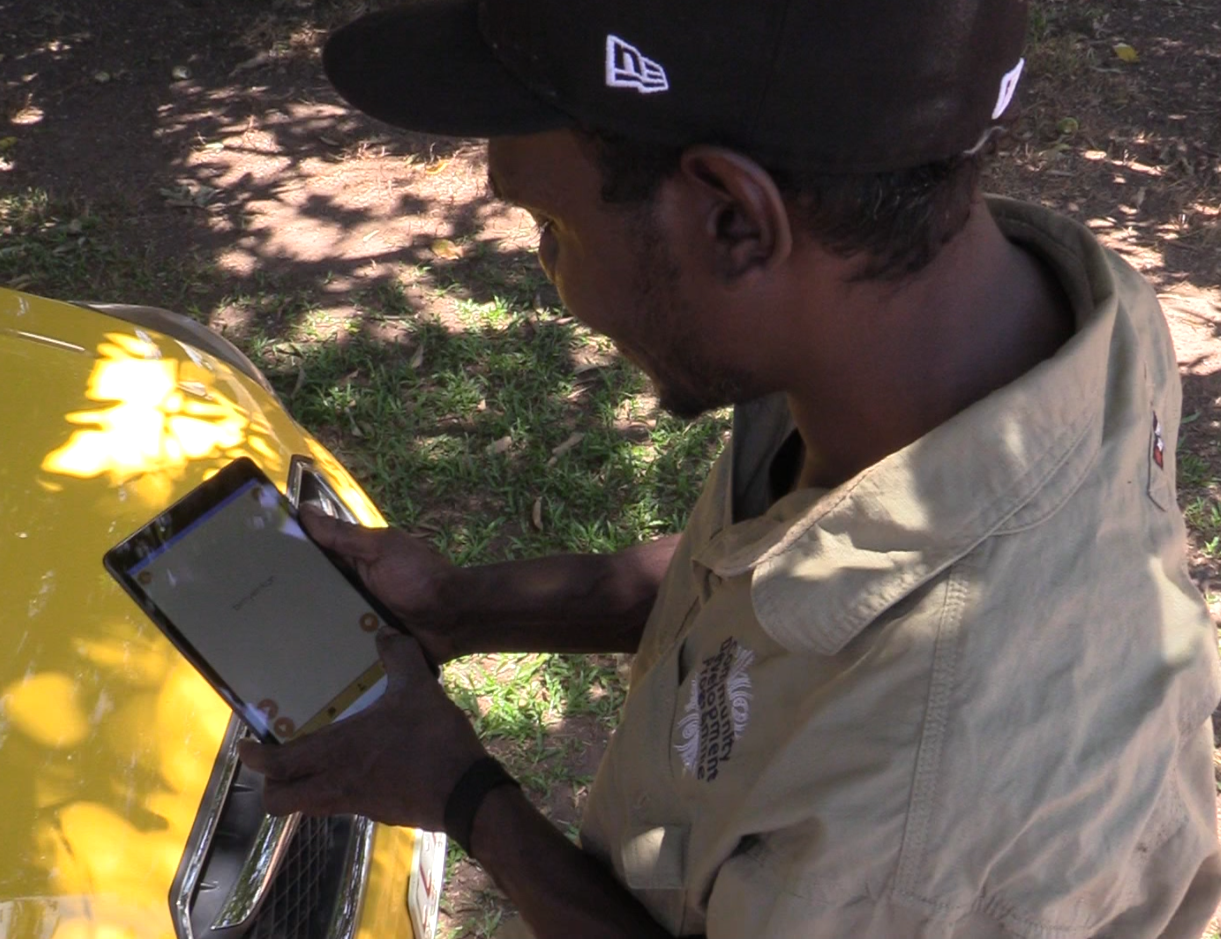} }}
    \caption{Lexical confirmation app}\label{fig:verif}
\end{figure}

Building on prior work developing mobile tools for language documentation \cite{bettinson2017developing}, we have begun to explore methods for deploying the pipeline in a remote community.
While the first step of identification of new words is straightforward, the task of lexical confirmation might be much more complex to apply.
The members of an Aboriginal community might not be familiar with technologies and they are not necessarily literate (in the narrow western sense).
Taking into account these constraints, we built a lexical confirmation app and trialled it on a small lexicon (Fig.~\ref{fig:verif}).

The idea would be to load the output of the spoken term detection system into the app.
Then a speaker can listen to the query,
listen to the utterance where we expect the query to be found,
and confirm if the utterance contains the query, (Fig.\ref{setup:2})


\section{Conclusion}

We investigated the use of spoken term detection methods as an alternative to the usual methods that have been inspired by automatic speech recognition,
and which require exhaustive transcription even for passages which exceed the present state of our knowledge about the language.
Instead we devised a workflow based on spoken term detection, and
simulated it for two small corpora: one in Mboshi, one in Kunwinjku.
The simulations of this workflow show that, with well chosen speech representations,
we may have a viable approach for rapidly bootstrapping transcriptions of large collections of speech
in endangered languages.
The next step of this work would be to design methods to involve Indigenous people in tasks such as the construction of the lexicon or the confirmation of the output of our system for a deployment of this workflow in a remote community.

\section*{Acknowledgements}

We are grateful to the Bininj people of Northern Australia for the opportunity to work in their community,
and particularly to artists at Injalak Arts and Craft (Gunbalanya) and to the Warddeken Rangers (Kabulwarnamyo).
Our thanks to several anonymous reviewers for helpful feedback on earlier versions of this paper.
The lexical confirmation app presented in this paper has been designed by Mat Bettinson, at Charles Darwin University.
This research was covered by a research permit from the Northern Land Council, ethics approved from CDU and was supported by the Australian government through a PhD scholarship, and grants from the Australian Research Council and the Indigenous Language and Arts Program.

\vfil\pagebreak
\renewcommand{\UrlFont}{\ttfamily\small}
\bibliographystyle{coling}
\bibliography{coling2020}

\begin{thebibliography}{}

\bibitem[\protect\citename{Bettinson and Bird}2017]{bettinson2017developing}
Mat Bettinson and Steven Bird.
\newblock 2017.
\newblock Developing a suite of mobile applications for collaborative language
  documentation.
\newblock In {\em Proceedings of the 2nd Workshop on the Use of Computational
  Methods in the Study of Endangered Languages}, pages 156--164.

\bibitem[\protect\citename{Bird}2020]{Bird20}
Steven Bird.
\newblock 2020.
\newblock Sparse transcription.
\newblock {\em Computational Linguistics}, 46(4).

\bibitem[\protect\citename{Boersma and Weenink}1996]{boersma1996praat}
Paul Boersma and David Weenink.
\newblock 1996.
\newblock Praat, a system for doing phonetics by computer, version 3.4.
\newblock {\em Institute of Phonetic Sciences of the University of Amsterdam,
  Report}, 132:182.

\bibitem[\protect\citename{Brinckmann}2009]{brinckmann2009transcription}
Caren Brinckmann.
\newblock 2009.
\newblock Transcription bottleneck of speech corpus exploitation.
\newblock {\em Proceedings of the 2nd Colloquium on Lesser Used Languages and
  Computer Linguistics}, pages 165 -- 179.

\bibitem[\protect\citename{{First Languages
  Australia}}2014]{australia2014angkety}
{First Languages Australia}.
\newblock 2014.
\newblock Angkety {M}ap: Digital resource report.
\newblock Technical report.
\newblock \url{https://www.firstlanguages.org.au/images/fla-angkety-map.pdf};
  accessed Nov 2020.

\bibitem[\protect\citename{Foley \bgroup et al.\egroup
  }2018]{foley2018building}
Ben Foley, Joshua~T Arnold, Rolando Coto-Solano, Gautier Durantin, T~Mark
  Ellison, Daan van Esch, Scott Heath, Frantisek Kratochvil, Zara
  Maxwell-Smith, David Nash, et~al.
\newblock 2018.
\newblock Building speech recognition systems for language documentation: The
  {CoEDL} {Endangered Language Pipeline and Inference System} ({ELPIS}).
\newblock In {\em Proceedings of The 6th International Workshop on Spoken
  Language Technologies for Under-Resourced Languages}, pages 205--209.

\bibitem[\protect\citename{Godard \bgroup et al.\egroup }2018]{godard2018very}
Pierre Godard, Gilles Adda, Martine Adda-Decker, Juan Benjumea, Laurent
  Besacier, Jamison Cooper-Leavitt, Guy-Noel Kouarata, Lori Lamel,
  H{\'e}l{\`e}ne Bonneau-Maynard, Markus Mueller, et~al.
\newblock 2018.
\newblock A very low resource language speech corpus for computational language
  documentation experiments.
\newblock In {\em Proceedings of the 11th International Conference on Language
  Resources and Evaluation}, pages 3366--70.

\bibitem[\protect\citename{Gupta and Boulianne}2020a]{gupta2020automatic}
Vishwa Gupta and Gilles Boulianne.
\newblock 2020a.
\newblock Automatic transcription challenges for {I}nuktitut, a low-resource
  polysynthetic language.
\newblock In {\em Proceedings of the 12th Language Resources and Evaluation
  Conference}, pages 2521--27.

\bibitem[\protect\citename{Gupta and Boulianne}2020b]{gupta2020speech}
Vishwa Gupta and Gilles Boulianne.
\newblock 2020b.
\newblock Speech transcription challenges for resource constrained indigenous
  language {C}ree.
\newblock In {\em Proceedings of the 1st Joint Workshop on Spoken Language
  Technologies for Under-Resourced Languages and Collaboration and Computing
  for Under-Resourced Languages (CCURL)}, pages 362--367.

\bibitem[\protect\citename{Jansen and Van~Durme}2011]{jansen2011efficient}
Aren Jansen and Benjamin Van~Durme.
\newblock 2011.
\newblock Efficient spoken term discovery using randomized algorithms.
\newblock In {\em 2011 IEEE Workshop on Automatic Speech Recognition \&
  Understanding}, pages 401--406. IEEE.

\bibitem[\protect\citename{Kamper \bgroup et al.\egroup
  }2015]{kamper2015unsupervised}
Herman Kamper, Micha Elsner, Aren Jansen, and Sharon Goldwater.
\newblock 2015.
\newblock Unsupervised neural network based feature extraction using weak
  top-down constraints.
\newblock In {\em 2015 IEEE International Conference on Acoustics, Speech and
  Signal Processing (ICASSP)}, pages 5818--22. IEEE.

\bibitem[\protect\citename{Kisler \bgroup et al.\egroup
  }2017]{kisler2017multilingual}
Thomas Kisler, Uwe Reichel, and Florian Schiel.
\newblock 2017.
\newblock Multilingual processing of speech via web services.
\newblock {\em Computer Speech and Language}, 45:326--347.

\bibitem[\protect\citename{Menon \bgroup et al.\egroup }2018a]{menon2018fast}
Raghav Menon, Herman Kamper, John Quinn, and Thomas Niesler.
\newblock 2018a.
\newblock Fast {ASR}-free and almost zero-resource keyword spotting using {DTW
  and CNNs} for humanitarian monitoring.
\newblock In {\em Proceedings of Interspeech 2018}, pages 2608--12.

\bibitem[\protect\citename{Menon \bgroup et al.\egroup }2018b]{Menon2018}
Raghav Menon, Herman Kamper, Emre Yilmaz, John Quinn, and Thomas Niesler.
\newblock 2018b.
\newblock {ASR-Free CNN-DTW} keyword spotting using multilingual bottleneck
  features for almost zero-resource languages.
\newblock In {\em Proceedings of the 6th International Workshop on Spoken
  Language Technologies for Under-Resourced Languages}, pages 182--186.

\bibitem[\protect\citename{Menon \bgroup et al.\egroup }2019]{menon2019feature}
Raghav Menon, Herman Kamper, Ewald van~der Westhuizen, John Quinn, and Thomas
  Niesler.
\newblock 2019.
\newblock Feature exploration for almost zero-resource {ASR-free} keyword
  spotting using a multilingual bottleneck extractor and correspondence
  autoencoders.
\newblock {\em Proceedings of Interspeech 2019}, pages 3475--3479.

\bibitem[\protect\citename{Park and Glass}2005]{park2005towards}
Alex Park and James~R Glass.
\newblock 2005.
\newblock Towards unsupervised pattern discovery in speech.
\newblock In {\em IEEE Workshop on Automatic Speech Recognition and
  Understanding}, pages 53--58. IEEE.

\bibitem[\protect\citename{Sakoe and Chiba}1978]{sakoe1978dynamic}
Hiroaki Sakoe and Seibi Chiba.
\newblock 1978.
\newblock Dynamic programming algorithm optimization for spoken word
  recognition.
\newblock {\em IEEE Transactions on Acoustics, Speech, and Signal Processing},
  26:43--49.

\bibitem[\protect\citename{Schneider \bgroup et al.\egroup
  }2019]{schneider2019wav2vec}
Steffen Schneider, Alexei Baevski, Ronan Collobert, and Michael Auli.
\newblock 2019.
\newblock Wav2vec: Unsupervised pre-training for speech recognition.
\newblock {\em Proceedings of Interspeech 2019}, pages 3465--69.

\bibitem[\protect\citename{Strand and Egeberg}2004]{strand2004cepstral}
Ole~Morten Strand and Andreas Egeberg.
\newblock 2004.
\newblock Cepstral mean and variance normalization in the model domain.
\newblock In {\em ISCA Tutorial and Research Workshop on Robustness Issues in
  Conversational Interaction}.

\bibitem[\protect\citename{Wittenburg \bgroup et al.\egroup
  }2006]{wittenburg2006elan}
Peter Wittenburg, Hennie Brugman, Albert Russel, Alex Klassmann, and Han
  Sloetjes.
\newblock 2006.
\newblock Elan: a professional framework for multimodality research.
\newblock In {\em 5th International Conference on Language Resources and
  Evaluation}, pages 1556--15.

\end{thebibliography}

\end{document}